\documentclass[10pt,letterpaper]{article}

\usepackage{cogsci}
\usepackage{graphicx}
\cogscifinalcopy 

\usepackage{pslatex}
\usepackage{apacite}
\usepackage{float} 

\usepackage{natbib}
\usepackage{hyperref}

\newcommand{\conceptARC}{ConceptARC}
\newcommand{\kidsarcsimple}{KidsARC-Simple}
\newcommand{\kidsarcconcept}{KidsARC-Concept}

\usepackage{xcolor}
\definecolor{BaseColor}{HTML}{708090} 
\definecolor{FineTunedColor}{HTML}{191970} 
\definecolor{MoEColor}{HTML}{006400} 
\definecolor{GPTColor}{HTML}{8B0000} 
\definecolor{HumanColor}{HTML}{B8860B} 

\usepackage{tikz}
\usepackage{makecell}
\usepackage{multirow}

\title{Do large language models solve ARC visual analogies like people do?}
 
\author{{\large \bf Gustaw Opiełka (g.j.opielka@uva.nl)}
  \AND {\large \bf Hannes Rosenbusch (h.rosenbusch@uva.nl)}
  \AND {\large \bf Veerle Vijverberg (v.p.vijverberg@uva.nl)}
  \AND {\large \bf Claire E. Stevenson (c.e.stevenson@uva.nl)} \\
  University of Amsterdam, Department of Psychological Methods, Amsterdam, Netherlands}

\begin{document}

\maketitle

\begin{abstract}
The Abstraction Reasoning Corpus (ARC) is a visual analogical reasoning test designed for humans and machines \citep{Chollet19}. We compared human and large language model (LLM) performance on a new child-friendly set of ARC items. Results show that both children and adults outperform most LLMs on these tasks. Error analysis revealed a similar "fallback" solution strategy in LLMs and young children, where part of the analogy is simply copied. In addition, we found two other error types, one based on seemingly grasping key concepts (e.g., Inside-Outside) and the other based on simple combinations of analogy input matrices. On the whole, "concept" errors were more common in humans, and "matrix" errors were more common in LLMs. This study sheds new light on LLM reasoning ability and the extent to which we can use error analyses and comparisons with human development to understand \textit{how} LLMs solve visual analogies.

\end{abstract}

\section{Introduction}
Until recently, visual analogy solving (e.g., \tikz \draw[fill=black] (0,0) circle (0.12cm) ; is to \tikz \draw (0,0) circle (0.12cm) ; as \tikz \draw[fill=black] (0,0) rectangle (0.24,0.24) ; is to ?) was considered something that is easy for humans, but out of reach for AI deep learning models \citep{mitchell2021}. However, large language models (LLMs) such as OpenAI's ChatGPT now appear capable of solving a range of analogy tasks in the text domain, including numeric and text-based versions of the matrix analogies in the Raven's Progressive Matrices \citep{webb2023emergent, hu2023rpmincontext} and, to a lesser degree, open-ended visual analogies \citep{Moskvichev23, mitchell2023arcgpt4, xu2023llmsarc}. The question then arises of \textit{how} LLMs solve these visual analogies. Is the process similar to adult humans that identify abstract relations and map these to new instances? Or perhaps more similar to the associative processes young children use? In this study, we compare human and LLMs visual analogy solving. More specifically, we use error analysis to understand \textit{how} LLMs obtain their solutions: is this through abstraction and analogy, association, or perhaps an entirely different process? 

Visual analogical reasoning can be assessed in both humans and AI models using the Abstraction Reasoning Corpus (ARC; Chollet, \citeyear{Chollet19}) and the ConceptARC \citep{Moskvichev23}. The ARC tasks are preferred above other visual reasoning tasks such as the Raven's Progressive Matrices \citep{Raven03} because these are open-ended rather than multiple-choice and can't easily be solved by chance, the open format also allows better tracing of human and LLM problem representations \citep{Johnson21} and the task is designed to assess numerous visual abstractions rather than a limited set of rules \citep{Chollet19, Moskvichev23}. However, current ARC tasks are too challenging for children as well as LLMs \citep{Moskvichev23, mitchell2023arcgpt4, xu2023llmsarc}. Therefore we created a small set of simplified ARC analogies, inspired by children's visual analogy tasks \citep[e.g.,][]{siegler2002microgenetic, hosenfeld1997indicators}, to gain insights into how LLMs' and children in various phases of analogical reasoning development solve visual analogies.

Young children's visual analogy solving is characterized by associative responses where one of the example images is (partly) duplicated or idiosyncratic responses are given (e.g., choosing a favorite shape and color) \citep{siegler2002microgenetic, hosenfeld1997indicators, stevenson2018pathsfigmatrix}. Generally between six and eight years (and younger with training), children start transitioning to successful analogy solving, marked by the appearance of partially correct solutions; here the underlying rule or concept is understood, but one or two aspects are missing \citep{stevenson2018pathsfigmatrix, hosenfeld1997indicators}. From 8-years onwards (or with more training) non-analogical responses disappear and (partial) analogical solutions take over \citep{stevenson2018pathsfigmatrix, hosenfeld1997indicators}. The transition from associative to analogical reasoning states in visual tasks coincides with what \citet{gentner1988} refers to as the relational shift. Interestingly, previous work comparing LLM and children analogical reasoning showed that LLMs tend to make similar associative errors as young children\citep{stevenson2023llmanalogy}. 

In the current study we compare LLM and human visual analogy solving, with a focus on exploring whether LLM errors resemble those of young children, adults or represent a less human-like process.

\section{Methods}
Data and code are available at \url{https://github.com/cstevenson-uva/kidsARC}.

\subsection{KidsARC-Simple and KidsARC-Concept}
We created two 8-item sets based on the ARC \citep{Chollet19} and \conceptARC{} \citep{Moskvichev23}. The Simple version is geared at younger children (4–8 years) and the Concept version at older children and adults (8+ years). 

\begin{figure}[t!]
\centering
\includegraphics[width=0.5\textwidth]{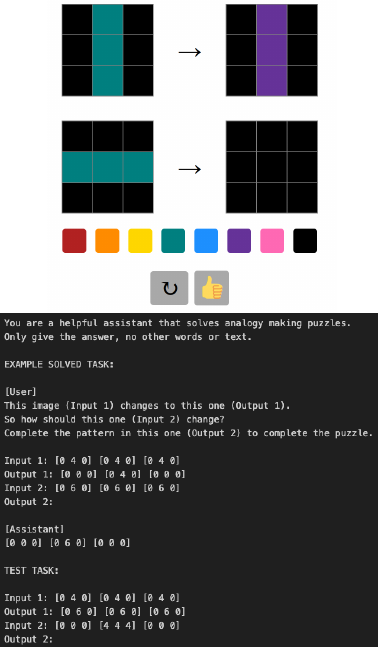}
\caption{Example item and interface from \kidsarcsimple{} task (top). Corresponding prompt given to LLMs (bottom), derived from \citep{Moskvichev23}.}
\label{fig:exampleitem}
\end{figure}

The original ARC and \conceptARC{} require few-shot learning with several input-output examples to derive the pattern and solve the item. We use only one input-output example, thereby mimicking the classical "A is to B as C is to D" analogy set-up, thus requiring one-shot learning (see Figure \ref{fig:exampleitem}). Having fewer input-output examples poses a challenge in creating unambiguous items, with some items having multiple solutions. However, multiple solutions provide a fruitful ground for testing distinct strategies used by humans and machines, which we discuss later.

\begin{figure}[h!]
\begin{center}
\includegraphics[width=0.4\textwidth]{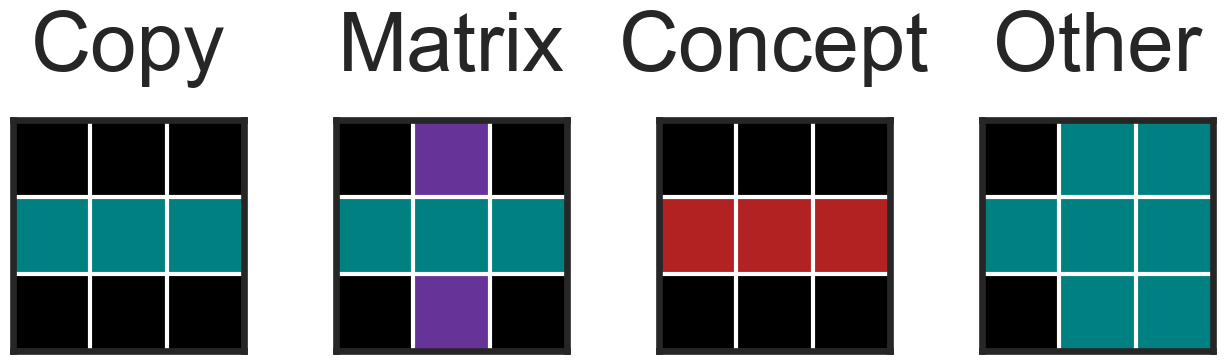}
\end{center}
\caption{Examples of different error types. See Figure \ref{fig:exampleitem} for the full item. A \textbf{Copy} error (partially) duplicates an input matrix. A \textbf{Matrix} error combines inputs. \textbf{Concept} errors apply correct transformations or concepts, but a mistake is made (here the wrong color). \textbf{Other} represents idiosyncratic responses.}
\label{fig:demo}
\end{figure}

\begin{figure*}[t!]
\begin{center}
\includegraphics[width=\textwidth]{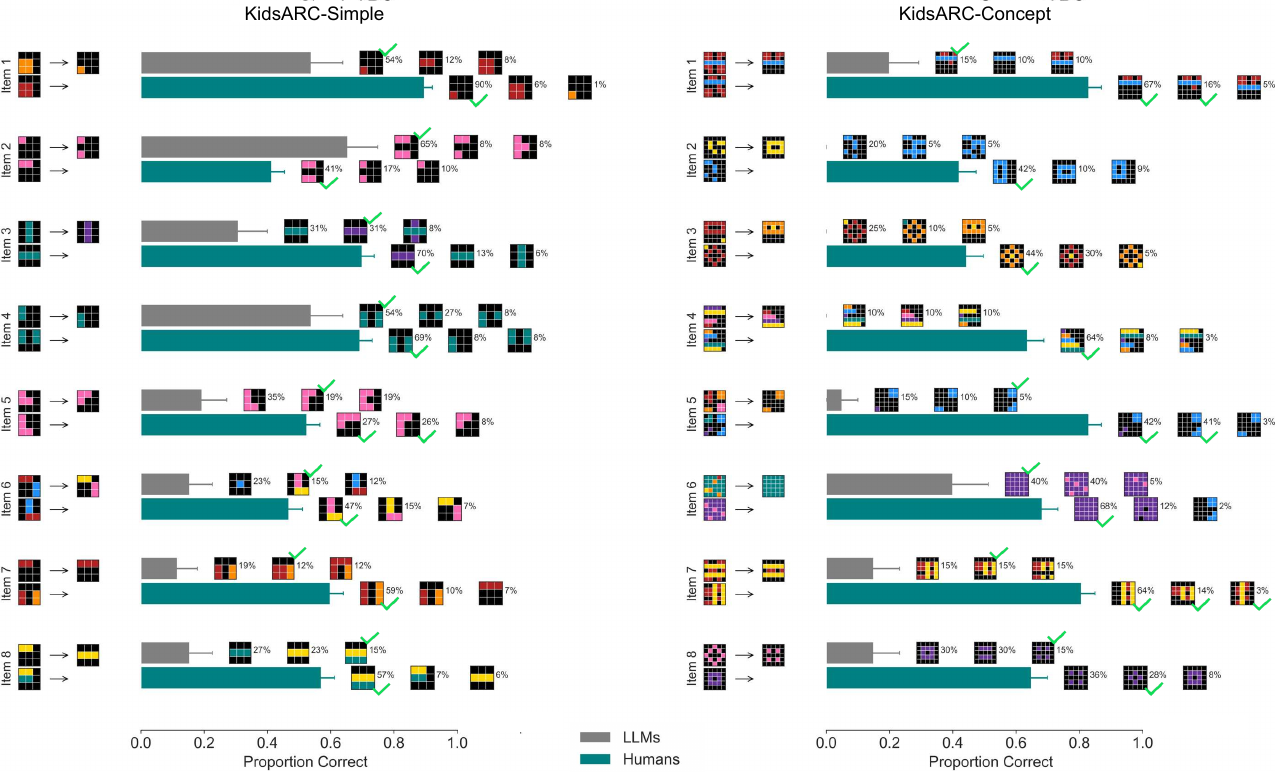}
\end{center}
\caption{Item-wise comparison of human vs. LLM performance on \kidsarcsimple{} and \kidsarcconcept{}. The items are visualized on the left of the performance bars. On the right we display the three most common responses (models and humans) along with percentage occurrence. Green ticks indicate correct responses. Note: items can have more than one correct response.}
\label{fig:performance}
\end{figure*}

\kidsarcsimple{} items consist of 3x3 grids representing simple concepts such as color and position. \kidsarcconcept{} consists of 5x5 matrices encoding concepts from the \conceptARC{}, such as, Inside-Outside and Complete Shape. See Figure \ref{fig:performance} for all items. 

\subsection{Human Data Collection}
Data collection took place at the NEMO science museum in Amsterdam. The room had several tables with a tablet that ran a cognitive task. Visitors, after signing informed consent, could solve one or more of the five tasks. Two of the tablets were used for this study. Participants received two examples. One to get familiarized with the user interface and one involving a simple analogy. Both examples had written verbal instructions that older children and adults could read and that we read out loud to younger children. All participants saw the items in the same order.

Due to the informal nature and noisy surroundings in the museum, verbal instructions were not always the same across participants. Also, although parents were requested to let their children solve tasks independently and participate themselves in different tasks, they still sometimes (implicitly) helped their children.

We collected data separately for the two item sets. Younger children who wanted to continue after the \kidsarcsimple{} were allowed to solve the \kidsarcconcept{} as well. Children solving both tasks were treated as separate participants. Both datasets had a positively skewed age distribution median of 8 (IQR = 7, 10) for \kidsarcsimple{} (\textit{n} = 155) and median of 12  (IQR = 9, 35) for the \kidsarcconcept{} (\textit{n} = 94). Overall, the youngest participant was 3, and the oldest 76. We binned participants into 5 age bins based on theory expectations: 3-5 (mostly non-analogical), 6-8 (transitioning), 9-11 (mostly analogical), 12+ (successful analogical reasoning) \citep{stevenson2018pathsfigmatrix}. 

\begin{figure*}[t]
\begin{center}
\includegraphics[width=\textwidth]{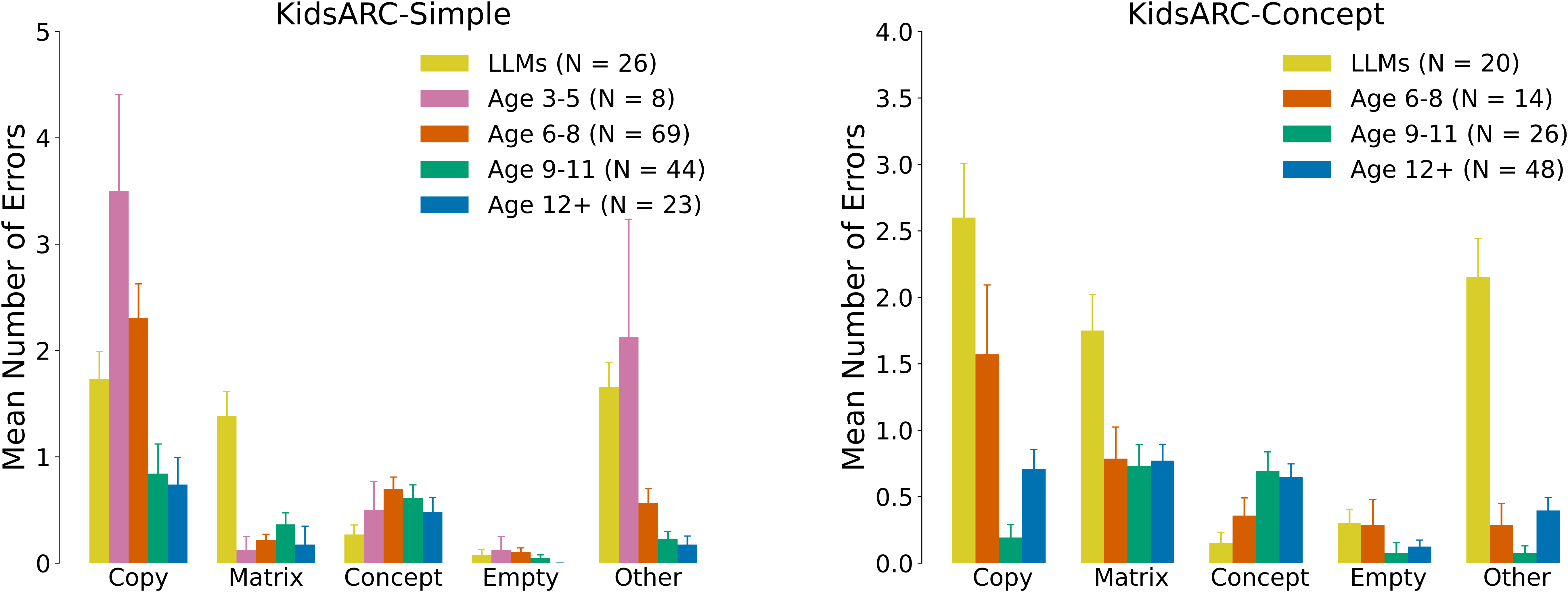}
\end{center}
\caption{As with young children, copying is a common error in LLMs. Compared to humans, matrix-based errors are more common in LLMs, while concept-based errors are less common. Other errors are common in both LLMs and young children.}
\label{fig:errors}
\end{figure*}

\begin{figure*}[t!]
\begin{center}
\includegraphics[width=\textwidth]{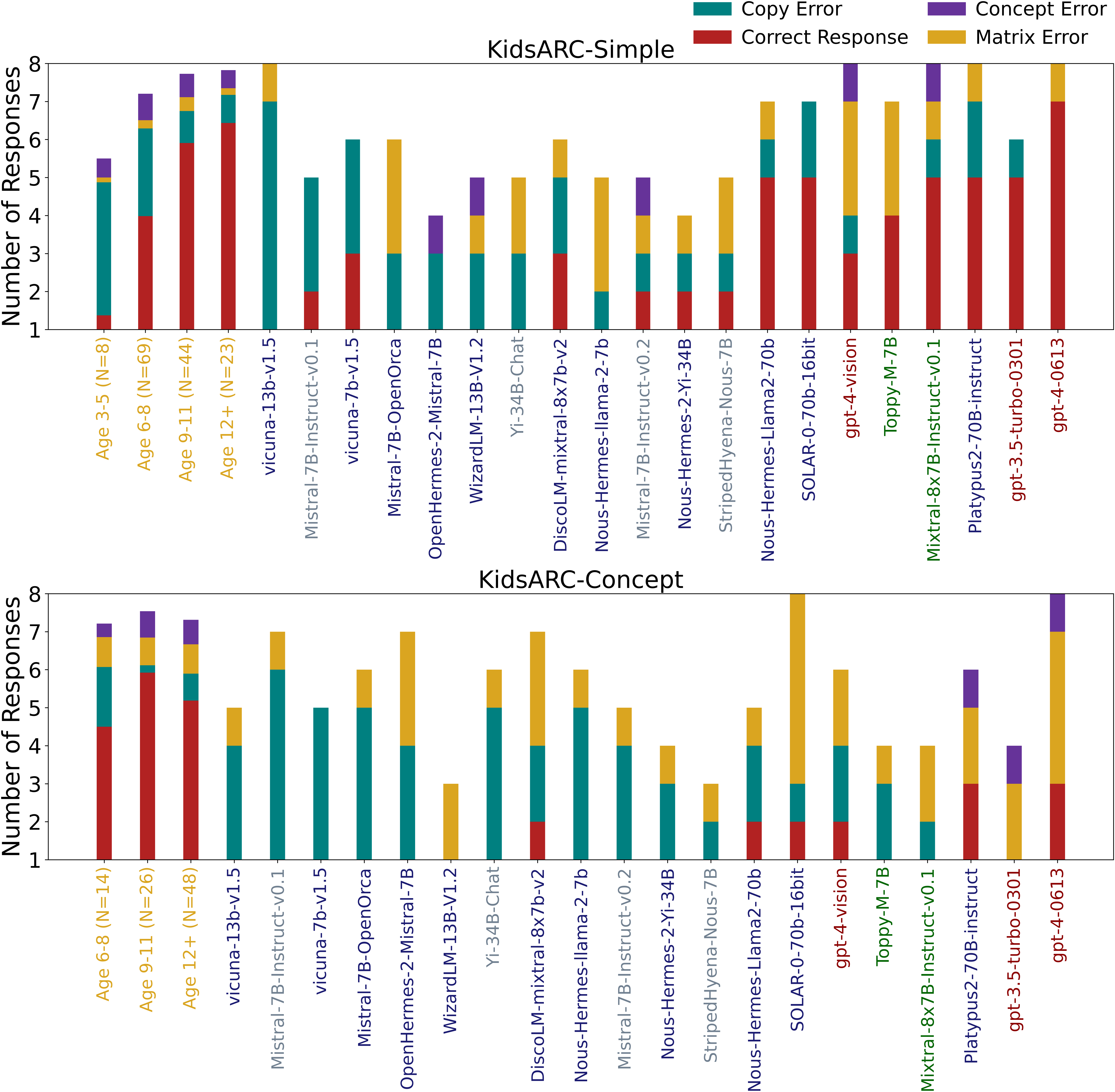}
\end{center}
\caption{Comparison of accuracy and error types model-by-model and for humans. We plot models with different training regimes - \textcolor{BaseColor}{Base Models}, \textcolor{FineTunedColor}{Fine-tuned Models}, \textcolor{MoEColor}{Mixture of Experts}, \textcolor{GPTColor}{GPTs}, as well as \textcolor{HumanColor}{Human participants}. \newline Note: To reduce clutter we do not plot 'other', empty, or invalid responses, hence some bars do not sum up to eight.}
\label{fig:spotlight}
\end{figure*}

\subsection{LLM Data Collection}
We used all 60 open-source LLMs available in the \href{www.together.ai/}{Together AI} collection. 20 of them failed to be called by the Together AI's API, leaving us with 40 models. Additionally, we used three GPT models from OpenAI: GPT-3 (01.03.2023 snapshot), GPT-4 (13.06.2023 snapshot), and the multimodal GPT-4 Vision (GPT-4V; preview version). 
Similarly to \citet{Moskvichev23}, we prompted the LLMs by putting the example matrices in text format and coding colors with numbers 0-9. Before presenting each test item, we included the same example item human participants saw, but solved. For GPT's the example item was included in the system prompt, and the test items in the user input. In the case of the LLMs accessed through the Together AI API, the example item was included in the same prompt as the current test item, since the API does not offer system prompt inputs. Similarly to human participants, LLMs were only allowed one attempt at the items and were given no feedback. The temperature was set to 0 for reproducible outputs for all LLMs.

To make use of the image processing capabilities of GPT-4 Vision, we also experimented with presenting the items visually and only requiring text for the output matrix (as with other LLMs) based on the numerical color codes. GPT-4V managed to follow the color coding and output format. However, it performed worse than when using the full text-based prompt that the other models received. Therefore, GPT-4V received the same prompt as other models and the item image was also included to further test its multimodal reasoning capabilities. The temperature setting was not available for GPT-4V at the time of access (20.01.2024).

\subsection{Exclusion Criteria}
Some exclusion criteria were specific to the LLMs, while some were shared across humans and models. We noticed that contrary to GPT models, many LLMs accessed from Together AI produced either incorrectly formatted responses or responses that were completely irrelevant to our items. Some of the common incorrect formatting included either duplicating the same response, providing more than one response, or providing a correctly formatted response followed by text irrelevant to the task. In such cases, we used regular expressions to extract whether a response contained correctly formatted output and if so, we treated the first output as the given response. Otherwise, we classified the responses as a no-response.

After pre-preprocessing LLM responses, we filtered out either human participants or individual LLMs if they had more than two no-responses in a given dataset. In human participants, this simply meant leaving the grid empty, and in the LLMs either a no-response, as defined above, or also an array full of black pixels. This left us with 144 human participants and 26 LLMs in the \kidsarcsimple{}, and 88 human participants and 20 LLMs in the \kidsarcconcept{}.

\subsection{Coding Erroneous Responses}
To examine differences in response strategies, we coded all erroneous responses for each item. Based on the literature on children's errors and an exploration of GPT-3's errors, we devised a taxonomy of three kinds of errors: (1) duplication, (2) concept-based and (3) matrix-based. Duplicating or (mostly\footnote{Some responses showed slight pixel changes. Nevertheless, exact copies constituted the majority (76\%) of the errors classified as copying.}) copying one of the analogy elements is the most common error response for children on visual matrix analogies \citep{stevenson2018pathsfigmatrix, hosenfeld1997indicators, siegler2002microgenetic}. Concept-based errors are partially correct solutions where the participant clearly understands the concept (e.g., position), but makes a mistake in its application (e.g., moving a pixel too far or forgetting to change the color); such errors are often seen in older children and on difficult items \citep{stevenson2018pathsfigmatrix, hosenfeld1997indicators}. Matrix-based errors are a new error type we encountered in this dataset; they are a result of simple matrix combinations of the analogy elements, e.g., the response includes all colored pixels from A, B and C. For example, see the third most common LLM response for item 3 in  \kidsarcsimple{} (Figure \ref{fig:performance}). The remaining responses were coded as "other" errors. Two independent raters, blinded to whether the response was made by humans or LLMs, coded each response into one of these four categories (initial agreement: 75\%), after which all discrepancies were resolved through discussion.

\section{Results}

\subsection{Aggregate Performance Differences}

In Figure \ref{fig:performance} we summarize the mean performance of all models and all human participants on both the \kidsarcsimple{} and \kidsarcconcept{} item sets. On the whole, humans perform better than the models with the exception of one item in the \kidsarcsimple{}. The difference in performance between the models and humans is even more pronounced when item difficulty increases. For example, on \kidsarcconcept{} an average LLM does not achieve human performance on any of the items. Also, quite strikingly, there are three items in the \kidsarcconcept{} that no model solves correctly. There is large variance in how individual LLMs performs, which we discuss in the next section. 

\begin{table}[H]
\begin{center} 
\caption{Comparison of Average Performance (\%) between children, adults, and LLMs.} 
\label{table} 
\vskip 0.12in
\begin{tabular}{llllll} 
\Xhline{2\arrayrulewidth}
 & \multicolumn{4}{c}{Age} & \multirow{2}{*}{LLMs} \\ 
\cline{2-5}
 & 3-5 & 6-8 & 9-11 & 12+ & \\
\cline{2-6}
\kidsarcsimple{} & 17.2 & 49.8 & 73.9 & 80.4 & 33.2 \\
\kidsarcconcept{} & \emph{NA} & 56.2 & 74.0 & 64.8 & 11.9\\
\Xhline{2\arrayrulewidth}
\end{tabular} 
\end{center} 
\end{table}

Figure \ref{fig:performance} also visualizes the individual items (left) and three most common responses for both the models and humans (next to their respective performance bars). Analyzing individual errors affords insights into the task-solving strategies employed by both humans and models. First, we see that both make the error of literally, or almost literally, copying one of the input matrices. For example, in \kidsarcsimple{} items 3 and 7 the most common response for the models and the second most common response for the humans was to copy the third input matrix (C-term). 

A new strategy we observed is what we call the 'matrix strategy', where the response is a result of some sort of pixel-level merging of the input matrices (see Methods section). This occurred far more often in LLMs. This strategy is well-illustrated by the third most common model responses in items 3 and 7 in the \kidsarcsimple{}, where the response can be conceptualized as the result of an element-wise union of the input matrices. Formally, if $A$, $B$, and $C$ are the input matrices, the response matrix $D$ is given by $D = A \lor B \lor C$, where $\lor$ denotes the element-wise logical OR operation. However, some of the responses that appear to be a result of a matrix strategy, cannot be easily formalized. E.g., in item 6 in the \kidsarcsimple{} the most common model response can be conceptualized as an XOR-like operation over the three input matrices (black pixels are coded as 0 in the experiment and all other colors are $\ge$ 1), except the 2\textsuperscript{nd} pixel in the 3\textsuperscript{rd} row (which would be red and not black according to the XOR). Finally, we also observed responses that were a result of direct matrix arithmetic, where LLMs treated the color codes as integers, e.g., in task 3 in 
\kidsarcsimple{} - $6 (\textrm{purple}) - 4 (\textrm{teal}) = 2 (\textrm{orange})$.

\subsection{Errors in LLMs and Humans}

In Figure \ref{fig:errors} we visualise how often, on average, LLMs and humans across age-groups produce different errors (copy, matrix-based, concept-based, empty responses, or other). We see that models primarily make copy, matrix, or other errors on both the \kidsarcsimple{} and \kidsarcconcept{} items. Similarly, for both younger "pre-analogical" children (3-5 year-olds) and 6-8 year-olds in "transitioning" phases, copying is the most common error and rapidly diminishes in older age-groups. "Other"/idiosyncratic errors are also common for the youngest age-group, but disappear in older groups. Matrix-based errors on the \kidsarcconcept{} items are more common in LLMs than humans. Concept errors, however, occur very rarely in the models compared to human respondents. The abundance of matrix solution strategies, and lack of concept-based errors shows that LLMs do not generalize abstract concepts as required by the ARC.

\subsection{Model Spotlight}

While there are general trends in LLM responses, there is still much variance in both performance and errors which we explore in Figure \ref{fig:spotlight}. Here we plotted the counts of different error types separately for each model. For easy viewing and comparison, we only include models that met the inclusion criteria in the \kidsarcconcept{}.

We make a couple of preliminary observations. First, there are models that primarily rely on duplication and/or matrix strategies, which are also the models that achieve the lowest performance. The models that perform the best are GPTs (which notably rarely use copying), Mixture of Experts models (although performance dropped in the \kidsarcconcept{}) and Platypus2-70B-instruct (which was the only model that performed on par with GPT-4 on \kidsarcconcept{}) . Interestingly, Platypus2-70B-instruct \citep{platypus2023} is a Llama-2 fine-tuned on a dataset involving \href{https://huggingface.co/datasets/garage-bAInd/Open-Platypus}{logical reasoning tasks} which is notable, since the base Llama-2 model performed so poorly that it did not even meet our inclusion criteria. ARC tasks were not part of the dataset.

Models that performed well on either of the tasks, generally had a high parameter count. This is excluding Mixture of Experts models, both having 7B parameters, 10 times less than similarly performing models. On the \kidsarcsimple{}, however, the smaller 7B parameter Vicuna model \citep{zheng_judging_2023} performed better than the bigger 13B version.

\section{Discussion}
In this paper, we compared how LLMs and children at different stages of analogical reasoning development, perform, and what kind of solution strategies they employed when solving ARC-like items. Our main findings show that LLMs are prone to (partially) copying the input matrices when giving a response - a fallback strategy that young children in pre-analogical and transitioning stages exhibit. What is more, we find that humans and LLMs differ in the types of errors they make. While humans make errors that are conceptually close to the correct solution but might miss a couple of pixels, LLMs often rely on simple combinations of the input matrices. We also found indications of how fine-tuning models on datasets designed to improve reasoning capabilities in LLMs might help in solving visual analogy tasks.

In our investigation, we not only gained insights by looking at error patterns but also recognized the value of including ambiguous items in our dataset. Specifically, two items from the \kidsarcconcept{} (items 1 and 7, shown in Figure \ref{fig:performance}) allowed for two valid solutions: applying conceptual knowledge related to spatial relationships (specifically, inside-outside concepts) or through a more straightforward approach of eliminating specific rows or columns. Notably, humans chose the conceptual approach 82\% (115/141) of the time, whereas LLMs did so only 14\% (1/7) of the time. While the small number of items and low LLM success rate limit strong conclusions, these findings highlight the value of designing ambiguous items for future research into AI and human analogical reasoning capabilities.

Our findings suggest that LLMs, while adept at learning surface statistics, fail to grasp underlying concepts, echoing longstanding critiques of connectionist systems by 
\citet{Fodor1988-FODCAC} and observed in modern AI \citep{greff_binding_2020}. Specifically, neural networks struggle with compositionality and objectness, particularly in tasks like ARC where object-based abstractions are needed for robust generalization \citep{xu_llms_2024}. Similarly, while neural networks are effective at statistical pattern matching, they fail at utilizing abstract structures, unlike humans \citep{kumar_disentangling_2023}.  Our results represent a behavioral example of LLMs' failure to develop symbol-level abstractions, leading to strategies that diverge from those used by humans.

There are a few limitations to consider. First, although we anticipate that measurement errors were mitigated across participants, replicating the experiment under controlled conditions is essential for confirming the reproducibility of our findings with human participants. Second, caution must be exercised when comparing humans to LLMs to avoid pitfalls such as assessing models under conditions dissimilar to those experienced by humans, as highlighted by \citet{ivanova_running_2023}. Despite efforts to standardize item presentation for both groups, inherent differences remain. For instance, presenting the ARC items as matrices to AI models as this and previous studies have done \citep[e.g., ][]{Johnson21, Moskvichev23}, may have inadvertently influenced LLMs to adopt simple matrix arithmetic strategies. However, this does not explain why GPT-4-Vision produced many matrix errors despite having access to the tasks in a visual format. Nevertheless, future research should aim to further align human and LLM task presentation.

Analogical reasoning is a cornerstone of human intelligence and creativity \citep{gentner_analogy_2017}. Embedding this capability into AI systems is crucial for achieving generalization capabilities required for robust and trustworthy AI-systems \citep{mitchell2021}. In human cognitive development, children exhibit various solution strategy phases in their journey towards proficient analogical reasoning. Our results show that currently LLMs are in the early stages of learning to solve visual analogies and show some non-human deviations.
\newpage

\section*{Acknowledgements}
This research was funded by the the Dutch Research Council (NWO) project "Learning to solve analogies: Why do children excel where AI models fail?" with project number 406.22.GO.029 awarded to Claire Stevenson. We thank Arseny Moskvichev and Melanie Mitchell for their valuable feedback on the children's adaptation of the ARC task and Han van der Maas for his insightful contributions to the project. We also thank, in random order - Ann Sophie Honikel, Jan-Luca Weigand, Milla Rosalina Pihlajamaki, Aaron Benjamin Lob, Leonidas Jung and Konradas Mikalauskas - for their attentiveness, persistence, and personal vibrance that implicitly or explicitly added great value to this work.

\bibliographystyle{apacite}

\setlength{\bibleftmargin}{.125in}
\setlength{\bibindent}{-\bibleftmargin}
\nocite{noauthor_togetherai_nodate}
\bibliography{CogSci_2024}

\end{document}